\documentclass[11pt,a4paper]{article}

\usepackage{tabularx}
\usepackage{INTERSPEECH2019}

\usepackage{xcolor}
\usepackage{times}
\usepackage{latexsym}
\usepackage{amsmath}
\usepackage{url}
\usepackage{algorithm}
\usepackage{amsfonts}
\usepackage[noend]{algpseudocode}
\usepackage{lipsum}
\usepackage{multirow}
\usepackage{framed}
\usepackage{graphicx}
\usepackage{hyperref}
\usepackage{longtable}
\usepackage [english]{babel}
\usepackage [autostyle, english = american]{csquotes}
\MakeOuterQuote{"}
\usepackage[font=scriptsize,labelfont=bf]{caption}

\newif\ifcomment\commenttrue
\usepackage{framed}
\usepackage{mdwlist}
\usepackage{latexsym}
\usepackage{xcolor}
\usepackage{nicefrac}
\usepackage{booktabs}
\usepackage{amsfonts}
\usepackage[T1]{fontenc}
\usepackage{bold-extra}
\usepackage{amsmath}
\usepackage{dsfont}
\usepackage{amssymb}
\usepackage{bm}
\usepackage{graphicx}
\usepackage{mathtools}
\usepackage{microtype}
\usepackage{multirow}
\usepackage{multicol}
\usepackage{latexsym,comment}

\newcommand{\gem}[1]{\mbox{\textsc{gem}}}
\newcommand{\abr}[1]{\textsc{#1}}

\newcommand{\explain}[2]{\underbrace{#2}_{\mbox{\footnotesize{#1}}}}

\newcommand{\g}{\, | \,}

\DeclareMathOperator*{\argmax}{arg\,max}

\newcommand{\emaillink}[1]{ {\small \href{mailto://#1}{\texttt{#1}}}}

\newcommand{\hidetext}[1]{}
\newcommand{\ignore}[1]{}

\ifcomment
\newcommand{\pinaforecomment}[3]{\colorbox{#1}{\parbox{.8\linewidth}{#2: #3}}}
\else
\newcommand{\pinaforecomment}[3]{}
\fi

\newcommand{\jbgcomment}[1]{\pinaforecomment{red}{JBG}{#1}}

\newcommand{\smallurl}[1]{ \begin{tiny}\url{#1}\end{tiny}}

\definecolor{lightblue}{HTML}{3cc7ea}
\definecolor{CUgold}{HTML}{CFB87C}
\definecolor{grey}{rgb}{0.95,0.95,0.95}
\definecolor{ceil}{rgb}{0.57, 0.63, 0.81}

\newcommand{\qb}[0]{Quizbowl}

\title{Mitigating Noisy Inputs for Question Answering}
\name{Denis Peskov,$^1$ Joe Barrow,$^1$ Pedro Rodriguez,$^1$ Graham Neubig,$^2$ Jordan Boyd-Graber$^3$}
\address{
	$^1$University of Maryland Department of Computer Science and \abr{umiacs}\\
	$^2$Carnegie Mellon University Language Technology Institute\\
        $^3$University of Maryland Department of Computer Science, iSchool, \abr{umiacs}, and \abr{lsc}}
\small \email{\{\href{mailto:dpeskov@cs.umd.edu}{dpeskov}, 
         \href{mailto:jdbarrow@cs.umd.edu}{jdbarrow}, 
         \href{mailto:pedro@cs.umd.edu}{pedro\}@cs.umd.edu}, 
         \emaillink{gneubig@cs.cmu.edu}, 
         \emaillink{jbg@umiacs.umd.edu}}
\date{}

\newcommand{\figfile}[1]{figures/#1}

\newcommand{\gn}[1]{}

\newcommand{\denp}[1]{}%\textcolor{red}{\small [#1 --DP]}}

\makeatletter
\def\BState{\State\hskip-\ALG@thistlm}

\makeatother

\begin{document}

\newcommand{\asr}{\textsc{asr}}
\newcommand{\dan}{\textsc{dan}}
\newcommand{\rnn}{\textsc{rnn}}
\newcommand{\unk}{<\textit{unk}>}

\maketitle

% abstract
\begin{abstract}

Natural language processing systems are often downstream of unreliable inputs: machine translation, optical character recognition, or speech recognition.
For instance, virtual assistants can only answer your questions after understanding your speech.
We investigate and mitigate the effects of noise from Automatic Speech Recognition systems on two factoid Question Answering (\textsc{qa}) tasks.
 Integrating confidences into the model and forced decoding of unknown words are empirically shown to improve the accuracy of downstream neural \textsc{qa} systems.  
We create and train models on a synthetic corpus of over 500,000 noisy sentences and evaluate on two human corpora from \qb{} and Jeopardy! competitions.\footnote{To appear at INTERSPEECH 2019}

\end{abstract}
%We propose four possible ways to integrate confidences into a Deep Averaging or Recurrent Neural Network and
%RESOLVED \jbgcomment{Abstract needs an additional sentence about the details of
%  the model; part of the abstract needs to singal to potential
 % reviewers what kind of models we're talking about.}

% intro
\section{Introduction}
\label{sec:introduction}

%TODO \prcomment{Need one more sentence here connecting back to general tasks}
%Noise from speech transcription makes Question Answering (\textsc{qa}) less reliable.  While most \textsc{qa} datasets focus on \emph{text}: e.g.,
% Would be nice to work back in, but its really not necessary for at least squad, tqa is more justifiable, but no asr data
% \textsc{sq}u\textsc{ad}~\cite{wang2017gated}, Trivia\textsc{qa}~\cite{joshi2017triviaqa}, \qb,
%We generate synthetic data as a weakly supervised approach to handle variation in human speakers.

Progress on question answering (\textsc{qa}) has claimed
 human-level accuracy.  However, most factoid \textsc{qa}
models are trained and evaluated on clean text input, which becomes
noisy when questions are spoken due to Automatic Speech Recognition
(\asr{}) errors.
This consideration is disregarded in trivia match-ups between machines
and humans: \textsc{ibm} Watson~\cite{Ferrucci10watson} on Jeopardy!
and \qb{} matches between machines and trivia
masters~\cite{Boyd-Graber:Feng:Rodriguez-2018} provide text data for
machines while humans listen.  A fair test would subject both humans
and machines to speech input.
%  We use the output of an \textsc{asr} system as input to our \textsc{qa} models.

Unfortunately, there are no large \textit{spoken} corpora of factoid
questions with which to train models; text-to-speech software can be used as a method for generating training data at scale for question answering models (Section~\ref{sec:data}).
Although synthetic data is less realistic than true human-spoken
questions it easier and cheaper to collect at scale, which is
important for training.  
These synthetic data are still useful; in
Section~\ref{sec:human-data}, models trained on synthetic data are
applied to human spoken data from \qb{} tournaments and Jeopardy!

\begin{figure}[t]
	\centering
	\includegraphics[width=\linewidth]{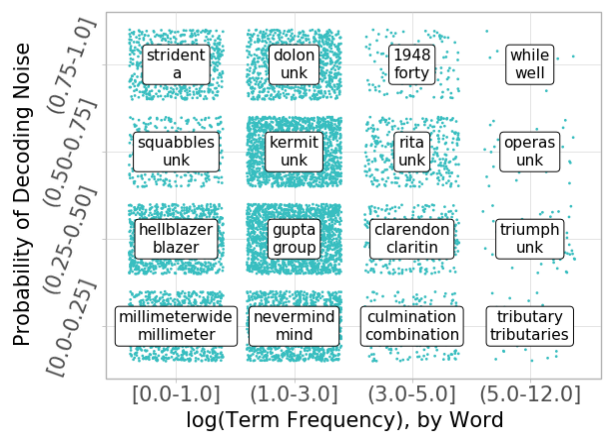}
	\caption{\textsc{asr} errors on \textsc{qa} data:
          original spoken words (top of box) are garbled (bottom).  While many words become
          into ``noise''---frequent words or the unknown
          token---consistent errors (e.g.,
          ``clarendon'' to ``clarintin'') can help downstream systems.  
          Additionally, words reduced to \unk{} (e.g., ``kermit'') can
          be useful through forced decoding into the closest
          incorrect word (e.g., ``hermit'' or even
          ``car'').}

\begin{comment}
\gn{There are lots of things I don't understand or think should be
fixed here. (1) What does the title ``Scoping Information Loss''
mean? Can it just be removed? Usually we don't have titles on
figures, in favor of captions. (2) What are the words in the 16
white boxes? Are they an example of one word out of the many that
belong to the class? (3) What are the Cyan dots in the back? (4) It
bugs me a little that your intervals are ``-0.001'' and ``1.001''
despite the fact that these values are inherently out of range. I'd
just use 0 and 1. (5) I don't understand why you focus on stopwords
here? Why not just calculate the probability of being
mis-recognized? (6) There is a mention in the main text of
``repetitive errors'', but I can't tell how we can tell errors are
repetitive by looking at this figure.}  \denp{Caption written by JBG
and sounds fine to me.  Frequency dots seem explainable.  Title and
small things can be changed if needed, but might be lower priority
than other tasks} \gn{Thanks! I think (2) has been resolved, and (1)
and (4) are low priority as you mentioned. (3) (5) and (6) are still
genuinely unclear though, so I'd try to improve the explanation.}
\jbgcomment{For CR, make sure to make the font bigger and remove the silly decimal places.}

%NOTRESOLVED.  Would be misleading/artificial.  NULL is non-mapping rather than mapped to unknown \jbgcomment{Changing NULL to unk (that you use later) would be good.}
\end{comment}
	\label{fig:noise_analysis}
\end{figure}

Noisy \asr{} is particularly challenging for \textsc{qa} systems
(Figure~\ref{fig:noise_analysis}).
While humans and computers might know the title of a ``revenge novel
centering on Edmund Dantes by Alexandre Dumas'', transcription errors
may mean deciphering ``novel centering on edmond dance by alexander
\unk{}'' instead.   Dantes and Dumas are low-frequency words in the English language and hence likely to be misinterpreted by a generic \asr{} model; however, they are particularly important for answering the question.  Additionally, the introduction of distracting words (e.g., ``dance'') causes \textsc{qa} models to make errors~\cite{jia-liang-2017-adversarial}.
Section~\ref{sec:noise} characterizes the signal in this noise: key
terms like named entities are often missing, which is
detrimental for \textsc{qa}.

% RESOLVED: \gn{As I noted, ``\textit{unk}'' is a rather unusual case. It'd be better to have an example without ``\textit{unk}s''.} \denp{unks are important to expansion section though so not sure if to remove.  slightly tweaked example}. \gn{Thanks! Looks better now.}
% \textsc{asr} introduces noise and challenges the clean-data assumption made by a typical \textsc{qa} model.

% I moved this, but it still seems out of place. Topic sentence should
% state basically: downstream systems lose confidence numbers

% More broadly, Natural Language Processing (\abr{nlp}) systems are
% often pipelines: downstream tasks depend on the output other
% \abr{nlp} systems.  A recommender system might use sentiment or
% topic models~\cite{blei2012probabilistic}, a parser might need to
% recover from errors to correct grammatical
% mistakes~\cite{sakaguchi2017error}, or a lattice machine translation
% system might use word segmentations~\cite{dyer2008generalizing}.
% Providing information about the accuracy or common errors of an
% upstream task can help the downstream system achieve its final goal.
% Most of the previous work on \textsc{qa} with \asr{} decouples the
% tasks.

%Previously proposed methods focusing on noisy \textsc{qa} include prioritizing machine learning training with a paragraph confidence metric \cite{clark2017simple}, handling noisy labels~\cite{natarajan2013learning}, and using long-term context to improve \textsc{asr} predictions~\cite{DBLP:journals/corr/abs-1802-02607}.
%Noise is also commonly addressed with Information Retrieval (\textsc{ir}) based approaches for tasks such as answering mobile queries~\cite{mishra2010qme}, building bots~\cite{leuski2009building}, and navigating knowledge bases~\cite{misu2007speech}.

Previous approaches to mitigate \asr{} noise for answering mobile
queries~\cite{mishra2010qme} or building bots~\cite{leuski2009building} typically use unsupervised methods, such as term-based information retrieval.
 Our datasets for training and evaluation can produce \textit{supervised} systems that directly answer spoken questions. Machine translation~\cite{sperber17emnlp} also uses \asr{} confidences; we evaluate similar methods on \textsc{qa}.

%RESOLVED \jbgcomment{A key criticism was the use of query expansion.  This is not clear a priori and also not a great fit for a non-IR venue.  Needs new name.}

Specifically, some accuracy loss from noisy inputs can be mitigated
through a combination of forcing unknown words to be decoded as the closest option
(Section~\ref{sec:forced-decoding}), and incorporating the uncertainties of
the \asr{} model directly in neural models
(Section~\ref{sec:conf-dan}).

The forced decoding method reconstructs missing terms by using terms similar to the transcribed input.
Word-level confidence scores incorporate uncertainty from the \asr{}
system into neural models.
Section~\ref{sec:exp} compares these methods against baseline methods
on our synthetic and human speech datasets for Jeopardy! and \qb{}.

\section{Spoken question answering datasets}
\label{sec:data}

%Query expansion and confidence models demonstrate an improvement over
%the baseline.  However, different lengths of questions and different
%neural architectures react differently.

%RESOLVED\jbgcomment{Jeopardy! questions are typically a single sentence, not  clear what ``self-contained'' is supposed to mean.}

%RESOLVEDgcomment{``ensure robustness'' is perhaps making too strong a claim.  describe the differences and let readers' draw their own conclusions}

%RESOLVED\jbgcomment{\qb kinda came out of nowhere, so would be good to introduce a little better when it makes sense; I updated to }

Neural networks require a large training corpus, but recording
hundreds of thousands of questions is not feasible. Crowd-sourcing
with the required quality control (speakers who say
``cyclohexane'' correctly) is expensive.
As an alternative, we generate a data-set with Google Text-to-Speech on 96,000 factoid questions from a trivia game called \qb{}~\cite{Boyd-Graber:Feng:Rodriguez-2018}, each with 4--6 sentences for a total of over 500,000 sentences.\footnote{\url{http://cloud.google.com/text-to-speech}}
We then decode these utterances using the Kaldi chain model~\cite{peddinti2015jhu}, trained on the Fischer-English dataset~\cite{cieri2004fisher} for consistency with past results on mitigating \asr{} errors in \textsc{mt}~\cite{sperber17emnlp}.  This model has a Word Error Rate (\textsc{wer}) of 15.60\% on the eval2000 test set.  The \textsc{wer} increases to 51.76\% on our \qb{} data, which contains out of domain vocabulary.   The most \textsc{bleu} improvement in machine translation under noisy conditions could be found in this middle  \textsc{wer} range, rather than in values below 20\% or above 80\%~\cite{sperber17emnlp}.  Retraining the model on the \qb{} domain would mitigate this noise;
however, in practice one is often at the mercy of a pre-trained
recognition model due to changes in vocabularies or speakers.
Intentional noise has been added to machine translation
data~\cite{michel2018mtnt, belinkov2017synthetic}.
Alternate methods for collecting large scale audio data include
Generative Adversarial Networks~\cite{donahue2018exploring} and manual
recording~\cite{lee2018odsqa}.

%RESOLVED when WER calculated  \jbgcomment{The substantive review that we got makes a good point
 % about WER.  Would be good to report this and to contrast this
  %dataset's WER with other comparable datasets.}

%RESOLVED\gn{This sentence is a bit of a suddenway to start the paragraph. Maybe just say: ``We test on two varieties of questions with different characteristics.''}
The task of \textsc{qa} requires the system to provide a correct answer out of many candidates based on the question's wording. We test on two varieties of different length and framing.  \qb{} questions, which are generally four to six sentences, tests a user's depth of knowledge; early clues are challenging and obscure but they progressively become easy and well-known.  Competitors can answer these types of questions at any point.
Computer \abr{qa} is competitive with the top players~\cite{yamada2018studio}.  Jeopardy! questions are single sentences and can only be answered after the question ends.  To test this alternate syntax, we use the same method of data generation on a dataset of over 200,000 Jeopardy questions~\cite{Dunn2017SearchQAAN}.

%RESOLVED\jbgcomment{This paragraph seems like it would make more sense when you introduce Kaldi}

%ISEDIT
%This drops for lattices to 20\% as all possible options are
%introduced.

%RESOLVED \jbgcomment{Not sure this is the right spot for model specifics; might be better with the model}

%ISEDIT
%\begin{figure}[t]
%	\includegraphics[width=.8\linewidth]{\figfile{asr_alignment.pdf}}
%	\caption{Certain words are lost by \textsc{asr} but a reasonable alignment with only small errors is possible with Giza++.}
%	\label{fig:alignment}
%\end{figure}

%\jbgcomment{While you talk about what is not a realistic goal, make sure you say what {\bf is}}

\subsection{Why \abr{qa} is challenging for \abr{asr}}
\label{sec:noise}

%\jbgcomment{Use \abr{tf-idf}.  Be precise about the the skyline}

%resolved\jbgcomment{This paragraph seems to be how ASR changes individual words and the vocabulary overall.  If that's so, then that should be the topic sentence: ``ASR changes the words in the dataset: the overall vocabulary is quite different, and important words are corrupted''.}

%Resolved \gn{This seems a bit scattershot and it's hard to tell what the main
  %take-aways should be. It might be good to say ``\asr{} changes the
  %features of the recognized text in sevearl important ways. (1) It
  %reduces the overall vocabulary. In our dataset .... (2) In
 % particular, it mis-recognizes important words. ..., (3) It deletes
  %words making the sentence length shorter. ...}

%RESOLVED\jbgcomment{Removed sentence about how 100\% accuracy is not goal: this seems out of place here; could go somehwere that's a better fit (e.g., when you introduce baselines on clean data):

% Achieving 100\% accuracy on this dataset is not a realistic goal, as
% not all test questions are answerable (specifically, some answers do
% not occur in the training data \jbgcomment{} and hence cannot be
% learned by a machine learning system).

%}

\asr{} changes the features of the recognized text in several important ways: the overall vocabulary is quite different and important words are corrupted.
First, it reduces the overall vocabulary.  In our dataset, the vocab drops from 263,271 in the original data to a mere 33,333.
This is expected, as \textsc{asr} only has 42,000 words in its vocab, so the long tail of the Zipf's curve is lost.
Second, unique words---which may be central to answering the question---are lost or misinterpreted; over 100,000 of the words in the original data occur only once.
Finally, \asr{} systems tend to deletes words which makes the sentences shorter; in our case, the average length decreases from 21.62 to 18.85 words per sentence.

The decoding system is able to express uncertainty by predicting \unk{}.
These account for slightly less than 10\% of all our word tokens, but is a top-2 prediction for 30\% of the 260,000 original words.
For \textsc{qa}, words with a high \textsc{tf-idf} measure are valuable.
While some words are lost, others can likely be recovered: ``hellblazer' becomes ``blazer'', ``clarendon'' becoming ``claritin''.
We evaluate this by fitting a \textsc{tf-idf} model on the Wikipedia dataset and then comparing the average \textsc{tf-idf} per sentence between the original and the \textsc{asr} data.  The average \textsc{tf-idf} score drops from 3.52 to 2.77 per sentence.

\section{Mitigating noise}
\label{sec:models}

%\jbgcomment{It seems that what are the training data have disappeared.  What trains the Jeopardy! data, for instance?}

%RESOLVED.  Aded sentence%\jbgcomment{A reviewer thought that the DAN didn't work.  Be more up front about what did and didn't work.}

%TODO\jbgcomment{Joe's comments are all good here: focus on what worked rather than a list of things tried.  Explain the IR baseline more clearly. }

%RESOLVED\jbgcomment{Section title should be more specific}

%TODO\jbgcomment{The introduction of IR is a little odd: you mention query expansion first before you talk about how it's an IR problem.  Other order is probably better.}

This section discusses two approaches to mitigating the effects of
missing and corrupted information caused by \textsc{asr} systems.  The
first approach---forced decoding---exploits systematic errors to arrive at the correct answer.
The second uses confidence information from the \asr{} system to
down-weight the influence of low-confidence terms.  Both approaches improve accuracy over a baseline \dan{} model and show promise for short single-sentence questions.   An \textsc{ir} approach is more effective on long questions.    

%TODO\prcomment{lapses in judgement seems like a weird/non-technical way to frame this: explicitly state what its doing}

\subsection{\textsc{ir} baseline}
The \textsc{ir} baseline reframes Jeopardy! and \qb~\textsc{qa} tasks as document retrieval ones with an inverted search index.  We create one document per distinct answer; each document has a text field formed by concatenating all questions with that answer together.  At test time questions are treated as queries, and documents are scored using \textsc{bm25}~\cite{ramos2003using,robertson2009probabilistic}.
We implement this baseline with Elastic Search and Apache Lucene.

\subsection{Forced decoding}
\label{sec:forced-decoding}

We have systematically lost information.  We could predict the answer if we had access to certain words in the original question and further postulate that wrong guesses are better than knowing that a word is unknown.
%We propose three forms of query expansion.

%RESOLVED BY REMOVAL OF ENTIRE PARAGRAPH \jbgcomment{Topic sentence
%not doing its job.  What in the data requires the alignment?  What
%does the alignment explain?  Frame it in that way: ``when ASR makes
%mistakes, it corrupts words\dots''}

%ISEDIT The first model uses IBM Model 3 to generate an alignment
%table between the corrupted \asr{} data and the original text data.
%We use our training data a second time and generate possible word
%candidates that were missed during decoding.  The second model uses a
%more robust version of the same information retrieval principle and
%looks at two-way relationships between \asr{} and original
%data~\cite{xu2008fire}.  Whereas the first model included many junk
%translations---stop-words such as ``unk'' or ``the'' would be mapped
%to a long tail of meaningful words---this version does not suffer
%from this problem: even if ``the'' maps to ``Monte'', ``Monte'' does
%not map back to ``the''.  These models had slightly inferior results
%and require extensive set-up compared to ur final appraoch.

%resolved\jbgcomment{Joe's suggestion is good.  Focus on a single
%model.  You can do ablation later but the narrative should only have
%one model}

%resolved \jbgcomment{This seems to be introduction of Kaldi, comes
%out of nowhere.  While you can probably assume that everyone know
%what it is, seems odd for it not to be formally introduced.}  ISEDIT
%The last model changes the logic of the \asr{} system itself.

We explore commerical solutions---Bing, Google,
\textsc{ibm}, Wit---with low transcription
errors.  However, their \textsc{api}s ensure that an end-user
often cannot extract anything more than one-best transcriptions, along
with an aggregate confidence for the sentence.  Additionally, the
proprietary systems are moving targets, harming reproducibility.

We use Kaldi~\cite{Povey11thekaldi} for all experiments.  Kaldi is a commonly-used, open-source tool for
 \textsc{asr}; its maximal transparency enables approaches that incorporate uncertainty into
downstream models.  Kaldi provides not only top-1
predictions, but also confidences of words, entire lattices, and phones
(Table~\ref{tab:data}).  Confidences are the same length as the
text, range from 0.0 to 1.0 in value, and correspond to the respective
word or phone in the sequence. 
%The mean one-best confidence for \qb{} data is 91\%.

\begin{table}[t!]
	\caption{As original data are translated through \abr{asr}, it
		degrades in quality.  One-best output captures per-word
		confidence.  Full lattices provide additional words and phone data captures the raw \abr{asr} sounds.  Confidence models and forced decoding could be used for such data.}
	\small
	\begin{tabularx}{.48\textwidth}{lXXr}
		\hline
		\\[-1em]
		%	\rotatebox{270}
		Clean  & For 10 points, name this revenge novel centering on Edmond Dantes, written by Alexandre Dumas \dots  \\ 
		\hline 
		\\[-1em]
		1-Best& for$^{0.935}$ ten$^{0.935}$ points$^{0.871}$ same$^{0.617}$ this$^{1}$ \ldots revenge novel centering on \unk{} written by alexander \unk{} \dots \\
		\hline 
		\\[-1em]
		``Lattice''& for$^{0.935}$ [eps]$^{0.064}$ pretend$^{0.001}$ ten$^{0.935}$  \dots \mbox{pretend}   point points  point   name same named name names this revenge novel \ldots \\
		\hline  
		\\[-1em]
		Phones  & f\_B$^{0.935}$ er\_E$^{0.935}$  t\_B$^{0.935}$  eh\_I$^{1}$  n\_E$^{0.935}$ \ldots p\_B   oy\_I n\_I t\_I s\_E sil s\_B ey\_I m\_E dh\_B ih\_I s\_E r\_B iy\_I v\_I eh\_I n\_I jh\_E n\_B aa\_I v\_I ah\_I l\_I \ldots \\
		
		\hline
	\end{tabularx}

	\label{tab:data}

\end{table}

The typical end-use of an \asr{} system wants to know when when a word is not recognized.  
By default, a graph will have a token that represents an unknown; in Kaldi, this becomes \unk{}.
At a human-level, one would want to know that an out of context word happened.

However, when the end-user is a downstream model,
a systematically wrong prediction may be better than a generic
statement of uncertainty.  So by removing all reference to \unk{} in
the model's Finite State Transducer, we force the system to decode
``Louis Vampas'' as ``Louisiana'' rather than \unk{}.  The risk we run
with this method is introducing words not present in the original
data.  For example, ``count'' and ``mount'' are similar in sound but
not in context embeddings.  Hence, we need a method to downweight
incorrect decoding.

\subsection{Confidence augmented \dan{}}
\label{sec:conf-dan}

% JOE:
%
% (C0) It feels like the "hinder sequence models" is an unsubstantiated claim
% that is only true if you're transfering a pretrained sequence model. I don't
% see why it would be necessarily true if you train a new sequence model from
% scratch. Perhaps make a syntactic claim?

%\jbgcomment{Topic sentence of this paragraph should explain why we use DAN.  }

%

We build on Deep Averaging Networks~\cite[\dan{}]{Iyyer:Manjunatha:Boyd-Graber:Daume-III-2015}, assuming
that deep bag-of-words models can improve predictions and be robust to
corrupted phrases.  The errors introduced by \textsc{asr} can hinder
sequence neural models as key phrases are potentially corrupted and
syntactic information is lost.

The original Deep Averaging Network, or \textsc{dan}, classifier has
three sections: a "neural-bag-of-words" (\textsc{nbow}) encoder, which
composes all the words in the document into a single vector by
averaging the word vectors; a series of hidden transformations, which
give the network depth and allow it to amplify small distinctions
between composed documents; and a softmax predictor.

The encoded representation~$\textbf{r}$ is the averaged embeddings of
input words. The word vectors exist in an embedding
matrix~$\textbf{E}$, from which we can look up a specific word~$w$
with $\textbf{E}[w]$. The length of the document is~$N$. To compute
the composed representation~$r$, the \textsc{dan} averages all of the
word embeddings:
%NOTRESOLVED\jbgcomment{I think this is wrong; isn't $N$ the length of the document?}  
\begin{equation}
\textbf{r} = \frac{\sum_{i}^{N}\textbf{E}[w\textsubscript{i}]}{N}
\end{equation}

The network weights~$\textbf{W}$, consist of a weight-bias pair for each layer of
transformations~$(\textbf{W\textsuperscript{(h\textsubscript{i})}, b\textsuperscript{(h\textsubscript{i})}})$ for each layer $i$ in the list of
layers~$L$. To compute the hidden representations for each layer, the
\textsc{dan}  linearly transforms the input and then applies a nonlinearity:
$
\textbf{h\textsubscript{0}} = \sigma (\textbf{W\textsuperscript{(h\textsubscript{0})}}\textbf{r}+\textbf{b\textsuperscript{(h\textsubscript{0})}})
$.
Successive hidden representations~$h\textsubscript{i}$ are:
$
\textbf{h\textsubscript{i}} = \sigma (\textbf{W\textsuperscript{(h\textsubscript{i})}}\textbf{h\textsubscript{i-1}}+\textbf{b\textsuperscript{(h\textsubscript{i})}})
$.
The final layer in the \textsc{dan} is a softmax output:
$
\textbf{o} = \mathrm{softmax}(\textbf{W\textsuperscript{(o)}}\textbf{h\textsubscript{L}} + \textbf{b\textsuperscript{(o)}})
$.
We modify the original \dan{} models to use word-level confidences from the \textsc{asr} system as a feature.  

%RESOLVED? \jbgcomment{``make use of'' is vague, be precise}

In increasing order of complexity, the variations are: a Confidence
Informed Softmax \textsc{dan}, a Confidence Weighted Average
\textsc{dan}, and a Word-Level Confidence \textsc{dan}.
We represent the confidences as a vector~$\textbf{c}$, where each cell
~$c\textsubscript{i}$ contains the \textsc{asr} confidence of word
$w\textsubscript{i}$.

% (C5) And then I would break out a small section to explain each of the models:
%ISEDIT
%\vspace{5mm} \noindent \textbf{Confidence Informed Softmax DAN} \\
%NOTRESOLVED but not worth effort\jbgcomment{make sure quotes are correct!  Also, this phrase is ideosyncratic to QB; good to find something more accessible} 

The simplest model averages the confidence across the whole sentence
and adds it as a feature to the final output classifier.  For example
in Table~\ref{tab:data}, ``for ten points'' averages to $0.914$. We introduce an additional weight in the output~$\textbf{W\textsuperscript{c}}$, which adjusts our prediction based on the average confidence of each word in the question.

%ISEDIT
\begin{comment}
We compute this confidence informed classification~$\textbf{o\textsuperscript{*}}$ as:

\begin{equation}
\textbf{o}^{*}  = \mathrm{softmax}([\textbf{W\textsuperscript{(c)}}; \textbf{W\textsuperscript{(o)}}][\frac{\sum_{i}^{N}c\textsubscript{i}}{N};\textbf{h\textsubscript{L}}] + \textbf{b\textsuperscript{(o)}})
\end{equation}

By concatenating the confidence weight~$\textbf{W\textsuperscript{c}}$ to the output weights and the averaged confidence to the final hidden representation.
\end{comment}

%\vspace{5mm} \noindent \textbf{Confidence Weighted Average DAN} \\

However, most words have high confidence, and thus the average confidence of a sentence or question level is high.  To focus on \emph{which} words
are uncertain we weight the word embeddings by their confidence attenuating uncertain words before calculating the \textsc{dan} average.

%ISEDIT

\begin{comment}
\begin{equation}
\textbf{r\textsuperscript{*}} = \frac{\sum \textbf{E}[w\textsubscript{i}] * c\textsubscript{i}}{N},
\end{equation}
\end{comment}

%\vspace{5mm} \noindent \textbf{Word-Level Confidence DAN} \\

Weighting by the confidence directly removes uncertain words, but this
is too blunt an instrument, and could end up erasing useful information contained in low-confidence words, so we instead learn a function based
on the raw confidence from our \abr{asr} system.  Thus, we recalibrate
the confidence through a learned function~$f$:

\begin{equation}
f(\textbf{c}) = \textbf{W\textsuperscript{(c)}c} + \textbf{b\textsuperscript{(c)} }
\end{equation}
and then use that scalar in the weighted mean of the \abr{dan}
representation layer:

\begin{equation}
\textbf{r\textsuperscript{**}} = \frac{\sum_{i}^{N} \textbf{E}[w\textsubscript{i}] * f(c\textsubscript{i})}{N}.
\end{equation}

In this model, we replace the original encoder~$\textbf{r}$ with the
new version $\textbf{r\textsuperscript{**}}$ to learn a transformation
of the \textsc{asr} confidence that down-weights uncertain words and
up-weights certain words.  This final model is referred to in the
results as ``Confidence Model''.

Architectural decisions are determined by hyperparameter sweeps.  They include: having a single hidden layer of 1000 dimensionality for the \dan, multiple drop-out, batch-norm layers, and a scheduled \textsc{adam} optimizer. Our \dan{} models train until convergence, as determined by early-stopping.  Code is
implemented in PyTorch~\cite{paszke2017automatic}, with TorchText for
batching.\footnote{Code, data, and additional analysis available at \smallurl{https://github.com/DenisPeskov/QBASR}}

\section{Results}
\label{sec:exp}

Achieving 100\% accuracy on this dataset is not a realistic goal, as
not all test questions are answerable (specifically, some answers do
not occur in the training data and hence cannot be
learned by a machine learning system).  Baselines for the \textsc{dan} (Table~\ref{table:combination_results}) establish realistic goals: a \textsc{dan} trained and evaluated on the \textit{same train and dev set}, only in the original
non-\textsc{asr} form, correctly predicts 54\% of the
answers. Noise drops this to 44\% with the best \textsc{ir} model and down
to $\approx30\%$  with neural approaches.

%% JBG 2019-06-29: This seems unfounded
% Hence, at the question-level, we believe that improvement up to 50\%
% might be possible with our methods. 
The noisy data quality makes full recovery unlikely and we view any
improvement over the neural model baselines as recovering valuable
information.  At the question-level, a strong \textsc{ir} model outperforms
the \textsc{dan} by around 10\%.  
Since \textsc{ir} can avoid all the noise while benefiting from additional independent data points, it
scales as the length of data increases.  There is additional motivation to investigate this task at the
sentence-level.  Computers can beat humans at the game by knowing certain questions immediately; the first sentence of the \qb{} question serves as a proxy for this threshold.  Our proposed combination of forced decoding with a neural model led to the highest test accuracy results and outperforms the \textsc{ir} one at the sentence level.
% We evaluate on the best confidence model, and the best expansion.

%RESOLVED \jbgcomment{remind reader about why multi-sentence is important (it's a quirk of one of our datasets).  Also make it clear which dataset you're talking about (or if it applies to both).}

A strong \textsc{tf-idf} \textsc{ir} model can top the best neural model at the multi-sentence question level in \qb{}; multiple sentences are important because they progressively become easier to answer in competitions.  However, our models improve accuracy on the shorter first-sentence level of the question.  This behavior is expected since textsc{ir} methods are explicitly designed to disregard noise and can pinpoint the handful of unique words in a long paragraph; conversely they are less accurate when they extract words from a single sentence.

%We achieve test accuracies of 0.269 and 0.248 on the baseline and Version 3 Expansion data respectively.  This is notably lower than our \dan{} and \rnn{}.  Integrating posterior confidences should lead to stronger results, but unlikely to overtake any of the proposed models here.

%Altering the Kaldi graph to never predict $unks$ and outputting potentially incorrect phones improved accuracy of the phone models from but

  %\jbgcomment{Make it easy for the reader to get what they need out of this table.  Some things that should be easy to see are:
	%\begin{itemize}
	%	\item How does the baseline IR and DAN system do on clean input?  (``Reference'' is unclear here\dots does that refer to implementation or data?)
%		\item How do these same systems do on the corrupted inputs (Where's IR?)
%		\item What's the best system this paper puts forward.  After Equation 3, it says that it's the ``Confidence'' model, but there's another system %below it.
	%\end{itemize}
	
%}

\begin{table}[t]

 %resolved  \jbgcomment{Make sure to use style checker generally.  ``lead to improvements'' is vague and verbose.  Favor verbs to nouns.}
		\caption{Both forced decoding (\textsc{fd}) and the best confidence model improve accuracy for the \dan{}.  Jeopardy only has an At-End-of-Sentence metric, as questions are one sentence in length.    Combining the two methods leads to a further joint improvement.  The \textsc{ir} and \textsc{dan} accuracies on clean data are provided as a reference.}
	\scriptsize
	\centering
	\begin{tabular}{ l c c c  c c c c}
		\toprule
		&& \multicolumn{2}{c}{\qb{}}  &&  \multicolumn{2}{c}{Jeopardy!}   \\
		\cmidrule(lr) {2-5}   	\cmidrule(lr) {6-7}
		&\multicolumn{2}{c}{Synth}& \multicolumn{2}{c}{Human} & Synth & Human\\
		\cmidrule(lr){2-3} \cmidrule(lr){4-5} \cmidrule(lr){6-6}  \cmidrule(lr){7-7}
		Method & \multicolumn{1}{c}{Start}&{End}& \multicolumn{1}{c}{Start}&{End} & & \\
		\cmidrule(lr) {1-7}
	    \multicolumn{6}{l}{\textbf{Methods Tested on Clean Data}}	\\
		\textsc{ir} & 0.064	& 0.544	& 0.400 & 1.000	 & 0.190 &0.050  \\
		\dan{} & 0.080 & 0.540 &0.200 & 1.000 &0.236 & 0.033 \\
		\cmidrule(lr) {1-7}
		\multicolumn{6}{l}{\textbf{Methods Tested on Corrupted Data}} \\
		\textsc{ir} & 0.021 & 0.442 & 0.180 & 0.560 & 0.079 & 0.050 \\
		\dan{} &  0.035 & 0.335 & 0.120 & 0.440  & 0.097 & 0.017  \\
		\textsc{fd}   & 0.032 & 0.354 & 0.120  & 0.440 & 0.102 & 0.033  \\
		Confidence  & 0.036 &  0.374 & 0.120  & 0.460 & 0.095  & 0.033 \\
		\textsc{fd}+Conf &  0.041  & 0.371  & 0.160 & 0.440  &  0.109 & 0.033  \\

		\bottomrule
	\end{tabular}

	\label{table:combination_results}
\end{table}

\subsection{Qualitative analysis \& human data}
\label{sec:human-data}
\begin{table}[t!]
	\caption{Variation in different speakers causes different transcriptions of a question on \underline{Oxford}, which can lead to different \dan{} predictions. }
	\scriptsize
	\setlength\tabcolsep{4pt}
	\begin{tabularx}{.50\textwidth}{p{1cm}p{7cm}}
	%	\begin{tabular}{l}
	\toprule
	 Speaker & Text \\
	 \midrule
	 Base & John Deydras, an insane man who claimed to be Edward II, stirred up trouble when he seized this city's Beaumont Palace.\\
	 \midrule
	 S1 & \unk{} an insane man who claimed to be the second \unk{} trouble when he sees \unk{} beaumont$\rightarrow$ \underline{The Rivals} \\
	  \midrule
	 %S2 & john dangerous insane man who claims to be the second stirring up trouble when he sees the city's beaumont $\rightarrow$ \underline{The Rivals}\\
	 S2 & \unk{} dangerous insane man who claim to be \unk{} second third of trouble when he sees the city's unk palace  $\rightarrow$ \underline{Rome}\\
	 \midrule
	 %ISEDIT
	S3 & \unk{} and then say man you claim to be the second stir up trouble when he sees the city's beaumont \unk{} $\rightarrow$ \underline{London}\\

	 %\midrule
	 %S3 &"unk and then say man you claim to be the second stir up trouble when he sees the city's beaumont unk",\\
	  %\midrule
	 %S4 & "unk dangerous insane man who claim to be unk second third of trouble when he sees the city's unk palace"\\
		%\rotatebox{270}{Clean}  & For 10 points, name this revenge novel centering on Edmond Dantes, written by %Alexandre Dumas.  \\
	\bottomrule
	\end{tabularx}

	\label{fig:human_data}
\end{table}

%RESOLVED eariler in intro rework \jbgcomment{Agree with Joe that this should be foregrounded more}

The synthetic dataset facilitates large-scale machine learning, but ultimately we care about performance on human data.
For \qb{} we record questions read by domain experts at a competition.  To account for variation in speech, we record five questions across ten different speakers, varying in gender and age; this set of fifty questions is used as the human test data.  Figure~\ref{fig:human_data} has examples of variations.  For Jeopardy! we manually parsed a complete episode by question.

\begin{comment}
\begin{table}
	\small
	\begin{tabularx}{.4\textwidth}{lXXr}
		\hline
		Model & Human Accuracy \\
		\hline
		Baseline DAN & XX \\
		\hline
		Query Expansion & XX  \\
		\hline
		Confidence DAN & XX \\
		\hline
		Lattices & XX \\
		\hline
	\end{tabularx}
	\caption{The relative ranks of the model accuracy on human data corresponds to the synthetic data accuracy.}
	\end{table}
\end{comment}

%\jbgcomment{``input about'' is vague.  Be direct: ``For a question on''}

The predictions of the regular \textsc{dan} and the confidence version
can differ.    For input about \underline{The House on Mango Street}, which contains
words like ``novel'', ``character'', and ``childhood'' alongside a
corrupted name of the author, the regular \textsc{dan} predicts
\underline{The Prime of Miss Jean Brodie}, while our version predicts
the correct answer.

%For a question on \underline{Pomp and Circumstance}, which
%contains words like ``composer'', ``marches'', and ``graduation'' with
%high confidences, the regular \textsc{dan} predicts Mahler's Symphony
%No 2, while our version predicts Edward Elgar, the composer.
%ISEDIT
%We visually analyze the embeddings between the models with t-SNE \cite{maaten2008visualizing}.

% discussion

\subsection{Discussion \& future work}
\label{sec:discussion}

%ISEDIT

%Recovery to perfect levels from noisy \textsc{asr} data does not
%appear to be possible, at least with a simple addition of
%confidences.  This in turn may mean that bespoke models are required
%for the similar tasks.  However, carefully scoping the problem for
%model feasibility is central to success.  We demonstrate that query
%expansion and end-to-end methods work better than more complicated
%approaches.

Confidences are a readily human-interpretable concept that may help build trust in the output of a system.
Transparency in the quality of up-stream content can lead to downstream improvements in a plethora of \textsc{nlp} tasks.

Exploring sequence models or alternate data representations may lead
to further improvement.  Including full lattices may mirror past results for machine
translation~\cite{sperber17emnlp} for the task of question answering.
Phone-level approaches work in Chinese~\cite{lee2018odsqa}, but our phone
models had lower accuracies than the baseline, perhaps due to a lack
of contextual representation.
Using unsupervised approaches for \asr{}
\cite{wessel2004unsupervised,lee2009unsupervised} and training \asr{}
models for decoding \qb{} or Jeopardy! words are avenues for further
exploration.

\section{Conclusion}
\label{sec:conclusion}

Question answering, like many \textsc{nlp} tasks are impaired by noisy inputs.
Introducing \asr{} into a \abr{qa} pipeline corrupts
the data.  A neural model that uses the \asr{} system's confidence outputs and systematic forced decoding of words rather than unknowns improves  \textsc{qa} accuracy on \qb{} and Jeopardy! questions.  
Our methods are task agnostic and can be applied to other supervised \textsc{nlp} tasks.
Larger human-recorded question datasets and alternate model approaches
would ensure spoken questions are answered accurately, allowing human
and computer trivia players to compete on an equal playing field.

%that challenge the robustness of \textsc{qa} downstream of \textsc{asr} of real speakers
%and explore models that can use contextual information to repair
%recognition errors.

%% the entire class
%% of speech-to-\textsc{nlp}-tasks research: machine
%% translation~\cite{sperber17emnlp}, syntatic
%% parsing~\cite{chappelier1999lattice}, dialogue state
%% tracking~\cite{henderson2014word}, and spoken language understanding.

%Predictions improve by similar margins between the models, which
%suggests extra information, rather than method of incorporation is
%paramount. Trainable embeddings neutralizes most gains from
%confidence information, suggesting that trained embeddings serve as
%de-facto confidences in current models.

\section{Acknowledgments}
\label{sec:ack}

This work was supported by NSF Grants IIS-1748663 and IIS-1748642.  The views expressed in this paper are our own.  We thank the reviewers, the \qb{} and Kaldi communities, and Yogarshi Vyas for their help.   

\bibliography{bib/journal-full,bib/denis}

\begin{thebibliography}{10}
\providecommand{\url}[1]{#1}
\csname url@samestyle\endcsname
\providecommand{\newblock}{\relax}
\providecommand{\bibinfo}[2]{#2}
\providecommand{\BIBentrySTDinterwordspacing}{\spaceskip=0pt\relax}
\providecommand{\BIBentryALTinterwordstretchfactor}{4}
\providecommand{\BIBentryALTinterwordspacing}{\spaceskip=\fontdimen2\font plus
\BIBentryALTinterwordstretchfactor\fontdimen3\font minus
  \fontdimen4\font\relax}
\providecommand{\BIBforeignlanguage}[2]{{%
\expandafter\ifx\csname l@#1\endcsname\relax
\typeout{** WARNING: IEEEtran.bst: No hyphenation pattern has been}%
\typeout{** loaded for the language `#1'. Using the pattern for}%
\typeout{** the default language instead.}%
\else
\language=\csname l@#1\endcsname
\fi
#2}}
\providecommand{\BIBdecl}{\relax}
\BIBdecl

\bibitem{Ferrucci10watson}
D.~A. Ferrucci, ``Build {W}atson: an overview of {DeepQA} for the {J}eopardy!
  challenge,'' in \emph{19th International Conference on Parallel Architecture
  and Compilation Techniques}, 2010, pp. 1--2.

\bibitem{Boyd-Graber:Feng:Rodriguez-2018}
J.~Boyd-Graber, S.~Feng, and P.~Rodriguez, \emph{Human-Computer Question
  Answering: The Case for Quizbowl}.\hskip 1em plus 0.5em minus 0.4em\relax
  Springer Verlag, 2018.

\bibitem{jia-liang-2017-adversarial}
R.~Jia and P.~Liang, ``Adversarial examples for evaluating reading
  comprehension systems,'' in \emph{Proceedings of Empirical Methods in Natural
  Language Processing}, 2017, pp. 2021--2031.

\bibitem{mishra2010qme}
T.~Mishra and S.~Bangalore, ``Qme!: A speech-based question-answering system on
  mobile devices,'' in \emph{Human Language Technologies:}, 2010, pp. 55--63.

\bibitem{leuski2009building}
A.~Leuski, R.~Patel, D.~Traum, and B.~Kennedy, ``Building effective question
  answering characters,'' in \emph{Proceedings of the Annual SIGDIAL Meeting on
  Discourse and Dialogue}, 2009, pp. 18--27.

\bibitem{sperber17emnlp}
M.~Sperber, G.~Neubig, J.~Niehues, and A.~Waibel, ``Neural lattice-to-sequence
  models for uncertain inputs,'' in \emph{Proceedings of the Association for
  Computational Linguistics}, 2017.

\bibitem{peddinti2015jhu}
V.~Peddinti, G.~Chen, V.~Manohar, T.~Ko, D.~Povey, and S.~Khudanpur, ``Jhu
  aspire system: Robust lvcsr with tdnns, ivector adaptation and rnn-lms,'' in
  \emph{Automatic Speech Recognition and Understanding (ASRU), IEEE Workshop
  on}, 2015, pp. 539--546.

\bibitem{cieri2004fisher}
C.~Cieri, D.~Miller, and K.~Walker, ``The fisher corpus: a resource for the
  next generations of speech-to-text,'' in \emph{Proceedings of the Language
  Resources and Evaluation Conference}, 2004.

\bibitem{michel2018mtnt}
P.~Michel and G.~Neubig, ``Mtnt: A testbed for machine translation of noisy
  text,'' in \emph{Proceedings of Empirical Methods in Natural Language
  Processing}, 2018.

\bibitem{belinkov2017synthetic}
Y.~Belinkov and Y.~Bisk, ``Synthetic and natural noise both break neural
  machine translation,'' in \emph{Proceedings of the International Conference
  on Learning Representations}, 2017.

\bibitem{donahue2018exploring}
C.~Donahue, B.~Li, and R.~Prabhavalkar, ``Exploring speech enhancement with
  generative adversarial networks for robust speech recognition,'' in
  \emph{IEEE International Conference on Acoustics, Speech and Signal
  Processing}, 2018, pp. 5024--5028.

\bibitem{lee2018odsqa}
C.-H. Lee, S.-M. Wang, H.-C. Chang, and H.-Y. Lee, ``Odsqa: Open-domain spoken
  question answering dataset,'' in \emph{2018 IEEE Spoken Language Technology
  Workshop (SLT)}.\hskip 1em plus 0.5em minus 0.4em\relax IEEE, 2018, pp.
  949--956.

\bibitem{yamada2018studio}
I.~Yamada, R.~Tamaki, H.~Shindo, and Y.~Takefuji, ``Studio {O}usia's quiz bowl
  question answering system,'' in \emph{NIPS Competition: Building Intelligent
  Systems}, 2018, pp. 181--194.

\bibitem{Dunn2017SearchQAAN}
M.~Dunn, L.~Sagun, M.~Higgins, V.~U. G{\"u}ney, V.~Cirik, and K.~Cho,
  ``Searchqa: A new {Q{\&}A} dataset augmented with context from a search
  engine,'' \emph{CoRR}, vol. abs/1704.05179, 2017.

\bibitem{ramos2003using}
J.~Ramos, ``Using tf-idf to determine word relevance in document queries,'' in
  \emph{Proceedings of the International Conference of Machine Learning}, 2003.

\bibitem{robertson2009probabilistic}
S.~Robertson, H.~Zaragoza \emph{et~al.}, ``The probabilistic relevance
  framework: Bm25 and beyond,'' \emph{Foundations and Trends in Information
  Retrieval}, vol.~3, no.~4, pp. 333--389, 2009.

\bibitem{Povey11thekaldi}
D.~Povey, A.~Ghoshal, G.~Boulianne, N.~Goel, M.~Hannemann, Y.~Qian, P.~Schwarz,
  and G.~Stemmer, ``The {K}aldi speech recognition toolkit,'' in \emph{IEEE
  Workshop on Automatic Speech Recognition and Understanding}, 2011.

\bibitem{Iyyer:Manjunatha:Boyd-Graber:Daume-III-2015}
M.~Iyyer, V.~Manjunatha, J.~Boyd-Graber, and H.~{Daum\'{e} III}, ``Deep
  unordered composition rivals syntactic methods for text classification,'' in
  \emph{Proceedings of the Association for Computational Linguistics}, 2015.

\bibitem{paszke2017automatic}
A.~Paszke, S.~Gross, S.~Chintala, G.~Chanan, E.~Yang, Z.~DeVito, Z.~Lin,
  A.~Desmaison, L.~Antiga, and A.~Lerer, ``Automatic differentiation in
  pytorch,'' in \emph{Conference on Neural Information Processing Systems:
  Autodiff Workshop: The Future of Gradient-based Machine Learning Software and
  Techniques}, 2017.

\bibitem{wessel2004unsupervised}
F.~Wessel and H.~Ney, ``Unsupervised training of acoustic models for large
  vocabulary continuous speech recognition,'' \emph{IEEE Transactions on Speech
  and Audio Processing}, vol.~13, no.~1, pp. 23--31, 2004.

\bibitem{lee2009unsupervised}
H.~Lee, P.~Pham, Y.~Largman, and A.~Y. Ng, ``Unsupervised feature learning for
  audio classification using convolutional deep belief networks,'' in
  \emph{Proceedings of Advances in Neural Information Processing Systems},
  2009, pp. 1096--1104.

\bibitem{papineni2002bleu}
K.~Papineni, S.~Roukos, T.~Ward, and W.-J. Zhu, ``{BLEU}: a method for
  automatic evaluation of machine translation,'' in \emph{Proceedings of the
  Association for Computational Linguistics}, 2002, pp. 311--318.

\bibitem{shannon1948mathematical}
C.~E. Shannon, ``A mathematical theory of communication,'' \emph{Bell system
  technical journal}, vol.~27, no.~3, pp. 379--423, 1948.

\end{thebibliography}
\bibliographystyle{IEEEtran}

\clearpage

\appendix

\section{Further Data Analysis}

One potential concern with the synthetically-generated dataset is that \asr{} systems might be either better or worse at recognizing text-to-speech(\textsc{tts}) speech.
If the \asr{}  system is trained on human data, then it might be an out-of-domain sample, or there might be systematic pronunciation issues that lower \asr{}  accuracy.  Alternatively, \textsc{tts}-generated speech might prove more regular or cleaner than human speech, so an \asr{} system may produce a higher transcription accuracy on this data.  Thus, we determine the distributional overlap between the \asr{} output  on both the synthetic and natural data.

\begin{figure}[t!]
	\begin{center}
	\includegraphics[width=\linewidth]{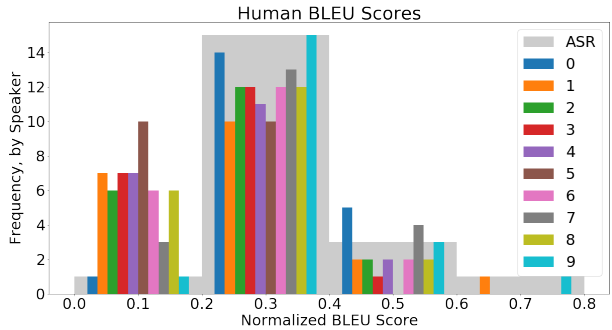}
    \caption{A comparison of \textsc{bleu} score distributions across human speakers (color-coded) to our artificial method, visualized by the step line.  The distributions of \textsc{bleu} scores are similar, with human data being slightly lower, justifying our weak supervision training approach.}

	\label{fig:bleu_variance}
	\end{center}
\end{figure}

\begin{figure}[t!]
	\begin{center}
		\includegraphics[width=\linewidth]{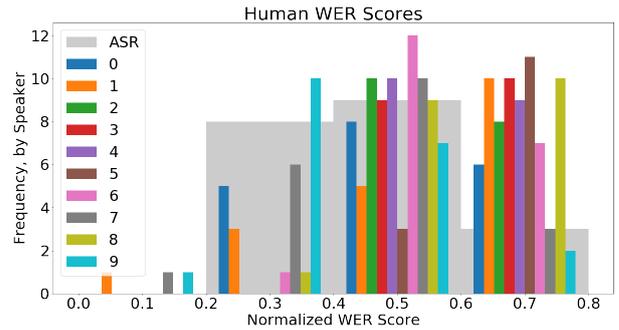}
		\caption{Similarly a comparison of \textsc{wer} score distributions across human speakers (color-coded) to our artificial method, visualized by the step line.  The distributions of \textsc{wer} scores are similar as well.  Speakers are color-coded.  The background step line is the \textsc{wer} of the automatic TTS approach.}
		
		\label{fig:wer_variance}
	\end{center}
\end{figure}

We compare \textsc{bleu} scores~\cite{papineni2002bleu} between the gold standard data and the decoded data for between the human and synthetic data variations.  By using \textsc{bleu} scores, which capture n-gram overlap between the target and source text, we can compare the variance in \asr{} between the two datasets.  Figure~\ref{fig:bleu_variance} illustrates this variance.  Additionally, Figure~\ref{fig:wer_variance} shows the comparison of Word Error Rate (\textsc{wer}).  Human data has more instances of higher \textsc{wer} and lower \textsc{bleu} scores than the auto-generated data on the same questions; however, the two sources of speech data generally follow a similar distribution and our results are comparable in accuracy to our synthetic data.  Therefore, we conclude that our method serves as a good approximation for the task, which allows weak supervision to work.

%\jbgcomment{I'm not sure the distribution comparison is particularly useful for main paper; could move to appendix.}

\section{Negative Results}

Alternative methods were applied to mitigate \asr{}-induced noise in the course of experimentation, including noisy channel techniques typically used in Information Retrieval and lattice-structured Recurrent Neural Networks.
For completeness, we discuss the results of these two experiments in this section.
While neither method provided an improvement on the question answering task, their discussion might prove useful for future research.

\subsection{Noisy Channel Expansion}

In both Information Retrieval and \textsc{nlp} it is often useful to model processes that induce noise using Shannon's noisy channel model~\cite{shannon1948mathematical}.
We know the answer would be predictable if we had access to certain words in the original question.
The noisy channel model allows us to reconstruct the original data as cleanly as possible by modeling the process by which noise was induced, in this case the trip from text to speech and back to text.
We propose two forms of query expansion based on this model, both of which are typically used in Cross Language Information Retrieval.

The first model uses IBM Model 3 to generate an alignment table between the corrupted \asr{}  data and the original text data.
The alignment table serves as the underlying corruption model which we are aiming to reverse.
We use our training data a second time and generate possible word candidates that were missed during decoding.

The second model uses a more robust version of the same Information Retrieval technique looks at two-way translations between \asr{} and original data based on (Xu, 2008).
Whereas the first model included many junk translations---stop-words such as ``unk'' or ``the'' would be mapped to a long tail of meaningful words---this version does not suffer from this problem: even if ``the'' maps to ``Monte'', ``Monte'' does not map back to ``the''.

In both cases, the reconstructed data was used to train the \textsc{DAN} model.
That neither was able to improve over the confidence modeling \textsc{DAN} indicates that the errors made by the \asr{} system were likely not recoverable with the translation models we used.
This is unsurprising, as many low-frequency important words were mapped to a handful of high-frequency terms, collapsing the space and preventing simple recoverability.

\subsection{Lattice-Structured RNN}

The confidence models are not calculate on a full lattice, and hence cannot not reconstruct alternate paths in situations with low confidences.
A more complex model can ingest the entire lattice, and not the top word prediction.
The lattice can update multiple words needed, as their relationships are preserved.
``Leo Patrick'' can now be reinterpreted as ``Cleopatra'', as the lattice relationship allows alternate paths to be explored.
The confidence values provide additional value about what path to follow within a lattice.

We produce three variations:
\begin{enumerate}
\item  A ``lattice'' \textsc{lstm}  that consumes the full lattice by linearizing the graphs with a topological sort and feeding it through a normal \textsc{lstm}.  
\item  A lattice \textsc{lstm}  without confidences.   This network only sees the word vectors when consuming the lattice structure.
\item  A lattice \textsc{lstm}  with confidences integrated as features.  The confidences are concatenated to the word vector inputs.
\end{enumerate}

This sequence demonstrates the gain from each part of the model.
The first tests the benefit of additional data.
The second tests the benefit of the structure of this data.
The third tests the importance of the confidence of each item in the data.

Unfortunately, none of these experiments outperformed the confidence augmented \textsc{DAN}.
These may be due to instability or training issues, however.

\end{document}